\documentclass[letterpaper, 10 pt, conference]{ieeeconf}  

\IEEEoverridecommandlockouts                              

\usepackage{graphics} 
\usepackage{epsfig} 
\usepackage{mathptmx} 
\usepackage{times} 
\usepackage{amsmath} 
\usepackage{amssymb}  
\usepackage{amsfonts}
\usepackage[cal=cm]{mathalfa}
\usepackage{lipsum}
\usepackage{cite}
\usepackage{multirow}

\usepackage[hidelinks]{hyperref}
\usepackage{algpseudocode}
\usepackage{xcolor}
\usepackage[ruled,vlined,linesnumbered]{algorithm2e}    

\title{\LARGE \bf
Interactive Navigation with Adaptive Non-prehensile Mobile Manipulation}

\author{Cunxi Dai$^{1*}$, Xiaohan Liu$^{1*}$, Koushil Sreenath$^{2}$, Zhongyu Li$^{2}$, Ralph Hollis$^{1}$
\thanks{*Equal contribution.}
\thanks{$^{1}$The Robotics Institute, Carnegie Mellon University, Pittsburgh, PA 15213, USA
        {\tt\small {cunxid,xiaohan5,rhollis}@cs.cmu.edu}}%
\thanks{$^{2}$University of California, Berkeley, CA, USA
        {\tt\small {zhongyu\_li,koushils}@berkeley.edu}}%
\thanks{This work was supported in part by NSF grant CNS-1629757 and Amazon. The work by Z. Li and K. Sreenath is supported by InnoHK of the Government of the Hong Kong Special Administrative Region via the Hong Kong Centre for Logistics Robotics.}%
}
\begin{document}

\maketitle
\thispagestyle{empty}
\pagestyle{empty}

\begin{abstract}
This paper introduces a framework for interactive navigation through adaptive non-prehensile mobile manipulation. A key challenge in this process is handling objects with unknown dynamics, which are difficult to infer from visual observation. To address this, we propose an adaptive dynamics model for common movable indoor objects via learned $SE(2)$ dynamics representations. This model is integrated into Model Predictive Path Integral (MPPI) control to guide the robot's interactions. Additionally, the learned dynamics help inform decision-making when navigating around objects that cannot be manipulated. Our approach is validated in both simulation and real-world scenarios, demonstrating its ability to accurately represent object dynamics and effectively manipulate various objects. We further highlight its success in the Navigation Among Movable Objects (NAMO) task by deploying the proposed framework on a dynamically balancing mobile robot, Shmoobot. Project website: \href{https://cmushmoobot.github.io/AdaptivePushing/}{\textcolor{blue}{https://cmushmoobot.github.io/AdaptivePushing/}}. 

\end{abstract}

\section{INTRODUCTION}

Interactive navigation refers to a navigation task where the robot interacts with the environment to successfully navigate, such as by moving obstacles that obstruct its path. This task is critical for robots operating in indoor environments, where numerous objects such as chairs, boxes, and other obstacles are frequently moved around by human activity. Non-prehensile mobile manipulation is an effective way for robots to interact with obstacles, particularly in tasks where the robot needs to reposition objects to make way for navigation. To enable this, there remain two main challenges that need to be addressed:

The first challenge is to adapt the manipulation strategy based on the object's dynamic properties such as mass, friction, and motion constraints. While some of these properties can be inferred from the visual classification of the object's category, relying solely on visual information may be insufficient when dealing with objects with irregular characteristics. For instance, the dynamics of a wheelchair, when one wheel has more friction than the other, a heavily loaded box, or a cart with locked wheels, can be challenging to identify only using vision. In such cases, accurately identifying the object's dynamic properties often requires physical interaction and observation of its motion, similar to the way humans assess objects. Furthermore, this is even more challenging for dynamically balancing robots due to the planning and control complexity for both the object and the robot\cite{6224884, 6224970, 10160507}.

The second challenge is to effectively utilize the identified object dynamics for planning in interactive navigation. For instance, if the obstructing object is not manipulable (e.g., it is too heavy or locked), the robot must plan a potentially more costly route around the obstacle. Conversely, if the object is manipulable, the robot should reposition it to a desirable location where it will not collide with the environment and also facilitate smooth navigation.
\begin{figure}[t!]
    \centering
    \includegraphics[width=\columnwidth]{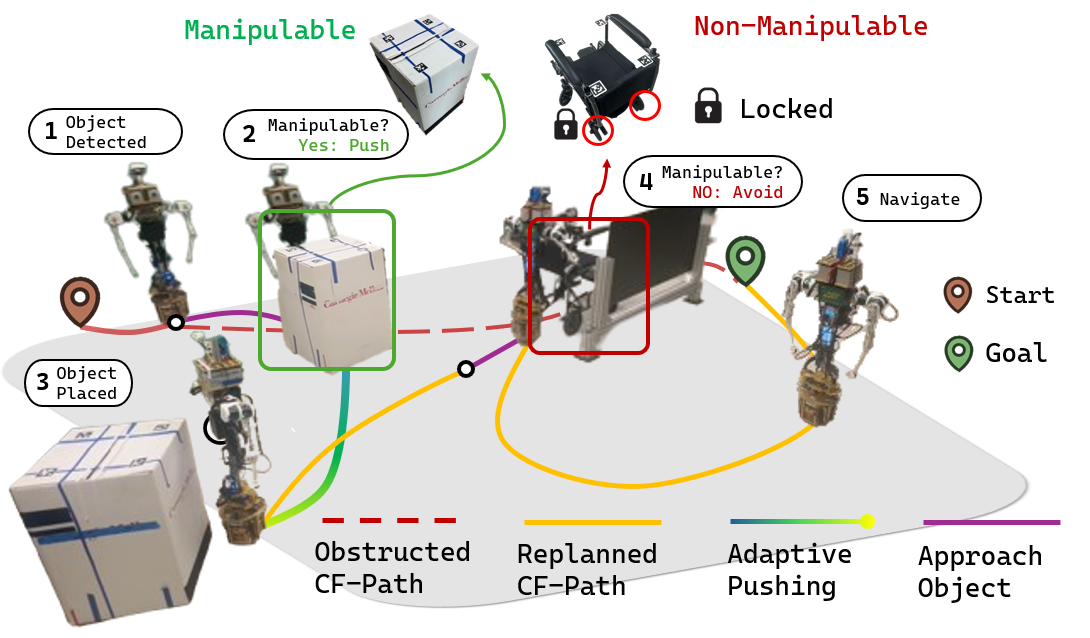}
    \caption{\textbf{Interactive Navigation with Adaptive Pushing:} Time-lapse of interactive navigation with adaptive pushing on Shmoobot, demonstrating the robot's ability to either push or plan a collision-free path (CF-Path) to navigate around an obstructing object based on its manipulability. The figure showcases two common object types: a light, manipulable box and a non-manipulable wheeled cart (e.g., a wheelchair with locked wheels).}
    \label{fig:introduction}
    \centering
    \vspace*{-3mm}
\end{figure}

We propose a framework that addresses these two challenges by utilizing an adaptive dynamics model for the object that enables non-prehensile manipulation and navigation decision-making. This framework achieves rapid adaptation to the object dynamics by leveraging a learned shared representation for common object dynamics in indoor settings. Objects such as boxes, shelves, carts, chairs, etc. are considered. We separated the learning for robot and object dynamics based on the observation that the robot dynamics remain consistent across different tasks, reducing the need for online adaptation. As a result, we only need to fine-tune the robot model with approximately 5 minutes of real-world data before deployment on the real robot, while directly utilizing object dynamics learned from simulations. Both the robot dynamics and object dynamics are integrated into a Model Predictive Path Integral Control (MPPI) method for effective planning and control for non-prehensile mobile manipulation~\cite{7989202}. Additionally, we use model prediction rollouts from MPPI to inform decision-making during interactive navigation tasks. Finally, we evaluated our framework using a new type of indoor service robot, named Shmoobot, that balances on a single spherical wheel.

\begin{figure*}[t!]
    \centering
    \includegraphics[width=2\columnwidth]{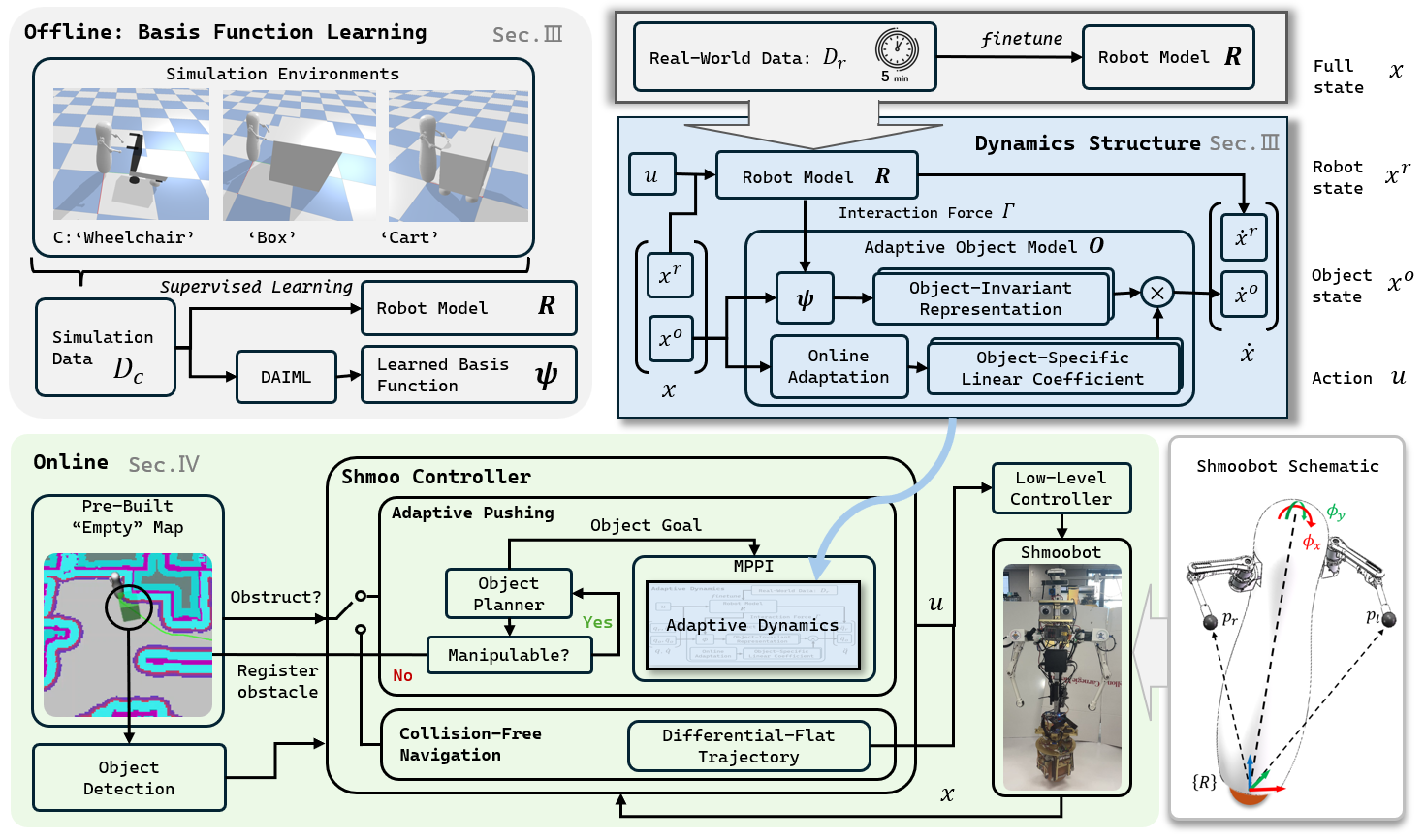}
    \caption{\textbf{Overall framework:}  In the offline phase, we first create simulation environments in Bullet~\cite{coumans2016pybullet} with three types of objects with randomized physical properties and train the basis function $\psi$ and robot model $R$. The learned $\psi$ are then used as parts of the adaptive dynamics, as described in Section.\ref{sec:AdaptivePushing}. In step 2, we utilize the adaptive dynamics to formulate the controller for interactive navigation. Given a pre-built map, the controller switches between adaptive pushing and collision-free navigation based on whether the object is manipulable or is blocking the way. The details of the interactive navigation framework are described in Sec.~\ref{sec:InteractiveNavigation}. Finally, the controller sends control actions to the low-level controller of Shmoobot as described in Sec.~\ref{sec:Experiment}. On the bottom-right corner, we show the schematic of Shmoobot and the necessary frames and notations.
 }
    \label{fig:control_diagram}
    \centering
    \vspace*{-3mm}
\end{figure*}

The main contributions of this paper are as follows:
\begin{enumerate}
\item We present a novel adaptive dynamics model that rapidly identifies object dynamics by leveraging a shared representation learned in simulation and directly applied to the real world. The framework accurately captures various planar object dynamics, as demonstrated through both simulation and real-world experiments.

\item We introduce a holistic interactive navigation framework that utilizes the model to reposition manipulable obstacles with unknown dynamics while avoiding non-manipulable ones during navigation.

\item The proposed framework realized novel applications on a dynamically balancing single-spherical-wheel robot (Shmoobot), allowing it to interactively navigate cluttered indoor environments with objects of unknown dynamic properties. The robot can clear a path by repositioning manipulable objects and skillfully navigate around non-manipulable ones.

\end{enumerate}


\section{Related Work\label{sec:RelatedWorks}}
Our work focuses on learning a shared representation for object dynamics to facilitate adaptive non-prehensile mobile manipulation and integration in interactive navigation. Here, we briefly discuss the fields related to our work. 
\subsection{Non-prehensile Mobile Manipulation}
In a mobile manipulation setting, non-prehensile manipulation such as pushing can be useful when the object is infeasible for grasping.  Several researchers used a statically stable mobile base and a rigid end-effector~\cite{9145591, Mericli}. In these cases, quasi-static assumptions are often posed due to the static nature of these tasks. However, for those cases using dynamic mobile manipulators~\cite{9387121, ferrolho2023roloma, 9034999, li2023kinodynamicsbasedposeoptimizationhumanoid, 8758922}, the robot's whole-body motion can be utilized for object manipulation and are optimized as part of the problem. For more complex and varying objects, previous work utilized system identification techniques~\cite{aguilera_control_2023, dai2024wheelchairmaneuveringsinglesphericalwheeledbalancing}. Such methods utilize an update law for the model parameters, and are limited to the expressiveness of the specified model.


\subsection{Learning Adaptive Dynamics}
Model-free methods typically train a network that aids the control policy by recovering environment parameters based on history I/O ~\cite{kumar2021rma,li2023robust}, or learning adaptive weights to achieve fast online-adaptation to new tasks~\cite{pmlr-v70-finn17a}. Although these methods proved to be robust, the identified parameter cannot be directly used for long-term decision-making. Other model-based approaches combine forward predictive models with an optimizer to evaluate multiple action parameters and execute an optimal action to achieve the desired goal~\cite{Kopicki}. Another line of research leverages representation learning, to acquire encoders for system inputs for interaction with the environment~\cite{O_Connell_2022}.

\subsection{Interactive Navigation}
Navigation among movable obstacle (NAMO) problems are studied to improve performance in cluttered environments, where only partial knowledge of objects is available. Previous research has primarily focused on geometry-based approaches, such as those in~ \cite{6631116, 7759546}. More recent studies have adopted learning-based methods, but these often rely on predefined skills to engage with objects~\cite{schoch2024insightinteractivenavigationsight, 8954627, khz2021interact}. One recent study updates object dynamics through physical interaction~ \cite{he2024interactivefarinteractivefastadaptablerouting}, yet still employs quasi-static assumptions, as seen in earlier works. These assumptions may be inadequate when object dynamics significantly influence the success of manipulation.


\section{Adaptive Pushing \label{sec:AdaptivePushing}}
The overall framework is illustrated in Fig.\ref{fig:control_diagram}. In this section, we first elaborate on how the adaptive dynamics model is structured, and then we discuss the data collection and learning process of the adaptive dynamics model and their integration into the adaptive pushing controller.

The robot platform we tested our method on is a dynamically balancing robot with a single spherical wheel called Shmoobot~\cite{lauwers_dynamically_2006, srivatchan_development_nodate}. We added a pair of 3-DoF arms to enable physical interaction with the environment. An advantage of Shmoobot is to utilize body-leaning to stably exert force on objects while moving~\cite{dai2024wheelchairmaneuveringsinglesphericalwheeledbalancing}. The low-level controller of Shmoobot includes the balancing controller and the arm controller. The balancing controller is a cascaded-PID controller that controls the body-leaning angle $\phi_x$ and $\phi_y$ with feed-forward Center-of-Mass (CoM) compensation for arm movements as discussed in~\cite{6224970} and~\cite{6225065}.  For the arm controllers, we adopt impedance control in Cartesian space and a steering controller to guide end-effector placement $ p_l^{des}, p_r^{des}$ w.r.t. robot frame $\left \{ R \right \} $ as shown in Fig.~\ref{fig:control_diagram}, thus enabling more stable pushing~\cite{dai2024wheelchairmaneuveringsinglesphericalwheeledbalancing}. The heuristic is formulated as $\left [ p_l^{des}, p_r^{des} \right ] =[x, y]^{\top} \pm R_{z}(\gamma)^{\top} l / 2$, where $x,y$ indicates the center of the object's contact surface, $R_{z}$ is rotation matrix on z-axis and $\gamma$ is the steering angle. Therefore, the controller input to the robot is $\mathbf{u}_t=\left [\phi_x,\phi_y, \gamma\right]$.

\subsection{Model Overview}
To enable a robot to manipulate objects with unknown dynamics, it's essential to account for both the dynamics of the robot and the object being manipulated. Given the full system state $\mathbf{x}_{t}$ at time $t$ partitioned as $\mathbf{x}_{t}=\left[\mathbf{q}_t,\dot{\mathbf{q}_t}\right]^{\top}$, where $\mathbf{q}_t$ and $\dot{\mathbf{q}_t}$ are the system configuration and its time-derivative, we seek a function $f$ so that the full state transition $F$ is:
\begin{equation}
    \mathbf{x}_{t+1}=F\left(\mathbf{x}_t, \mathbf{u}_t\right)=\left[\begin{array}{c}
\mathbf{q}_t+\dot{\mathbf{q}}_t \Delta t \\
\dot{\mathbf{q}}_t+f\left(\mathbf{x}_t, \mathbf{u}_t;c\right) \Delta t
\end{array}\right],
\end{equation}
where $\Delta t$ is the discrete-time increment, and $c$ is the object dynamics condition. Since the robot's dynamics remain consistent across different manipulation tasks, we decouple the learning of the robot's dynamics model $R$ from the object's dynamics model $O$ and only perform adaptation on $O$. The following elaborates on the details of this structure.
\subsection{Adaptive Dynamics Model Structure}
 For ease of elaboration on the model structure with two models, we introduce another partition of full state $\mathbf{x}_t$ as $\mathbf{x}_t=\left[\mathbf{x}_t^r,\mathbf{x}_t^o\right]^{\top}$, where the robot state is $\mathbf{x}_t^r=\left[\mathbf{q}^r_t,\mathbf{\dot{q}}^r_t\right]^{\top}$ and the object state $\mathbf{x}_t^o=\left[\mathbf{q}^o_t,\mathbf{\dot{q}}^o_t\right]^{\top}$. We formulate $f$ as a combination of $R$ and $O$ that predicts $\mathbf{\ddot{q}}^r_t$ and $\mathbf{\ddot{q}}^o_t$, the second-order derivative of robot and object configuration separately:
\begin{subequations}
\begin{align}
f\left(\mathbf{x}_t, \mathbf{u}_t;c\right) =\begin{bmatrix}
\mathbf{\ddot{q}}^r_t \\
\mathbf{\ddot{q}}^o_t
\end{bmatrix}=\begin{bmatrix}
R(\mathbf{x}_t^r,\mathbf{u}_t)_{\left[1\right]} \\
O(\mathbf{x}_t^o, \Gamma^o ;c)
\end{bmatrix}, \label{Za}\\
R(\mathbf{x}_t^r,\mathbf{u}_t)=
\begin{bmatrix}
 \mathbf{\ddot{q}}^r_t \\
\Gamma^r
\end{bmatrix},
\quad O(\mathbf{x}_t^o,\Gamma^o;c )=\mathbf{\ddot{q}}^r_o,\label{Zb}
\end{align}
\end{subequations}where $\Gamma^r$ and $\Gamma^o$ are the interaction force between the robot end-effector and the object represented in the robot and object frame. We represent $R$ with a multi-layer perceptron (MLP) and $O$ with a cross product of an object-invariant shared representation $\psi$ and object-specific linear coefficient $\mathbf{a(c)}$:
\begin{equation}
\label{eq:psia}
O(\mathbf{x}_t^o,\Gamma^o;c)\approx \psi\left ( \mathbf{x}_t^o,\Gamma^o\right ) \mathbf{a}(c),
\end{equation} 
where $O\in SE(2)$ since we focus on the planar dynamics of the objects, $\psi$ is a deep neural network (DNN) learned from simulation, and $\mathbf{a}(c)$ is a 1-dimensional vector conditioned on object dynamics condition $c$. This structure has been proven to be able to represent $O$ with arbitrary precision given that $\psi$ has enough neurons~\cite{O_Connell_2022}.

\subsection{Data Collection}

\subsubsection{Simulation}
We collected time-sampled data in simulation with human teleoperation over three different categories over common indoor objects (box, cart, wheelchair) with randomized dynamic characteristics (mass, dynamic constraints, friction, etc.) as $\mathcal{D}_c=\left\{\left(\mathbf{x}_t, \mathbf{\dot{x}}_t, \mathbf{u}_t, \Gamma\right)\mid t=1, \ldots, T\right\}$, where $\Gamma$ is the interaction force between the robot's end-effector and object in the world frame ${W}$ acquired from simulation. We can then calculate $\Gamma^r={_{\left\{W\right\}}^{\left\{R\right\}}}T\Gamma$ and $\Gamma^o={_{\left\{W\right\}}^{\left\{O\right\}}}T\Gamma$, where $\left\{R\right\}$ is the robot frame defined in Fig.\ref{fig:control_diagram},  $\left\{O\right\}$ is object frame at the object center,${_{\left\{W\right\}}^{\left\{R\right\}}}T$ and ${_{\left\{W\right\}}^{\left\{O\right\}}}$ are the transformation matrices between these frames. We collect time-stamped data $\left[\mathbf{q}_t,\mathbf{\dot{q}}_t,\mathbf{u}_t \right]$ and compute the acceleration $\mathbf{\ddot{q}}_t$ by numerical differentiation. 

\subsection{Learning Adaptive Dynamics}
\subsubsection{Adaptive Object Dynamics}
Objects have various dynamic properties that affect their motion. For example, wheelchairs typically have non-holonomic constraints constraining lateral movement while the cart box can be moved omnidirectionally.  We aim to learn a shared representation that encodes the similarities in these unique dynamics and leverage this to make online adaptation faster. To learn $\psi$, we adopt a Domain Adversarial Invariant Meta-Learning (DAIML) framework proposed by O’Connell et~al. for quadcopters~\cite{O_Connell_2022}, which utilizes an adversarial learning framework to overcome domain shift and learns an invariant representation among different conditions. To apply DAIML to object dynamics prediction, we modified it to use multi-step loss that provides better long-term prediction~\cite{lutter2021learningdynamicsmodelsmodel, hafner2019learninglatentdynamicsplanning,
Venkatraman2015ImprovingMP}. The multi-step loss function $\mathcal{L}_{multistep} $ is:
\begin{subequations}
\begin{align}
\mathcal{L}_{multistep}(\mathbf{y}_{c},\mathbf{x}^o_{c}) =\frac{1}{H}\sum_{h \in H} 
            E^{\top} \Sigma_{\mathcal{C}}^{-1}
            E, \\
E = \left( \mathbf{y}_{c}^{h} - \psi\left(\mathbf{x}_{c}^{o,h}, \Gamma^{o,h}\right) \mathbf{a}(c) \right),
\end{align}
\end{subequations}
where $H$ is length of steps used to calculate the loss, $\mathbf{y}_c$ is the ground truth data and $\Sigma_{\mathcal{C}}$ is the covariance matrix of dataset $\mathcal{C}$. During training of DAIML, the optimal linear coefficient $\mathbf{a}^{*}$ is calculated as:
\begin{equation}
    \mathbf{a}^{*}(c)=\arg \min _{a} \sum_{i \in B^{a}}\left\|\mathbf{y}_{c}^{i}-\psi\left(\mathbf{x}_{c}^{o,i}, \Gamma^{o,i}\right) \mathbf{a}(c)\right\|^{2},
\end{equation}
where $B^{a}$ is the data batch used to calculate the coefficient $\mathbf{a}^{*}$. Then the final training loss $\mathcal{L}^o$ is formulated as:

\begin{equation}
    \begin{split}
        \mathcal{L}^o=\max\limits_{h} \min\limits_{\psi, a_1, \dots, a_c} 
        \sum_{c \in C} 
        \sum_{i \in c} 
        ( 
            \operatorname{\mathcal{L}_{multistep} }(\mathbf{y}_{c}^i,\mathbf{x}_{c}^i) \\
            - \alpha \cdot \mathcal{L}_{CE} 
            ( h(\psi(x_{c}^{o(i)}, \Gamma^{o(i)})), c )
        )
    \end{split},
\end{equation}
where $h$ is another DNN that works as a discriminator to predict the object condition index $c$ out of $C$, $\alpha$ is the regularization ratio, and $\mathcal{L}_{CE}$ is the cross-entropy loss. 

\begin{figure}[t!]
    \centering
    \includegraphics[width=\columnwidth]{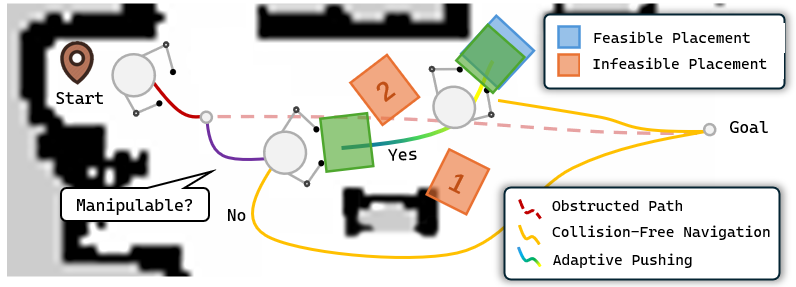}
    \caption{\textbf{Interactive Navigation Schematic:} First, the robot approaches the object that obstructs the planned path, followed by adaptive pushing towards the desired placement generated by the object planner. Some infeasible placement positions are shown respectively; 1: collision with the map and 2: collision with the planned path. If the object planner concludes that the object is not manipulable, the object is registered as an obstacle in the map and the robot re-plans with the updated map.}
    \label{fig:ineractive navigation}
    \centering
    \vspace*{-3mm}
\end{figure}

\begin{figure*}[t!]
    \centering
    \includegraphics[width=2\columnwidth]{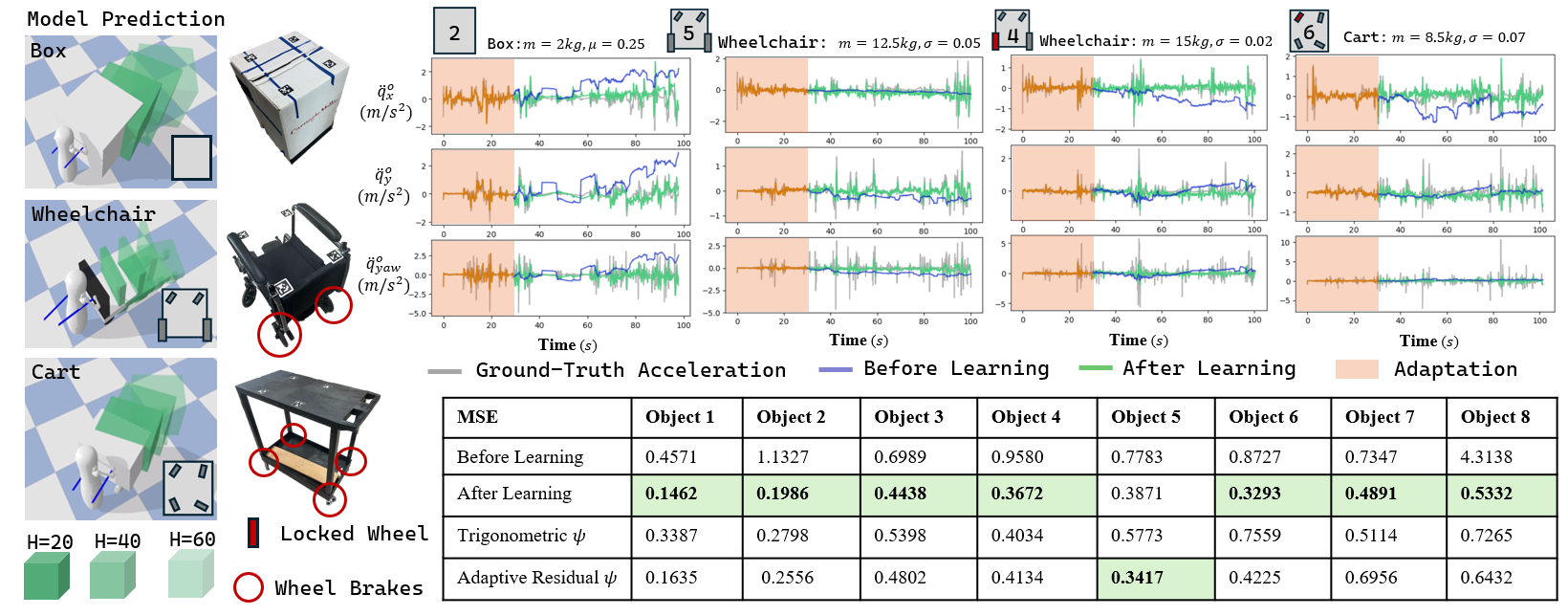}
    \caption{\textbf{Adaptive Dynamics Accuracy:} We compared the accuracy of the adaptive dynamics model $O$ prediction before and after learning on 8 objects whose dynamics were not seen during training. ($m$ is the object's mass, $\mu$ is the planar friction, $\sigma$ is the wheel's rotational friction and the red block on the data icon indicates a locked wheel). The left side is a visualization of the model roll-out of the different horizons, the interaction force is visualized in blue. We further compared with other models discussed in Sec.~\ref{sec:Validation} of which we listed the prediction MSE at the bottom.
}
    \label{fig:modelprediction}
    \centering
    \vspace*{-3mm}
\end{figure*}

\begin{figure}[t!]
    \centering
    \includegraphics[width=\columnwidth]{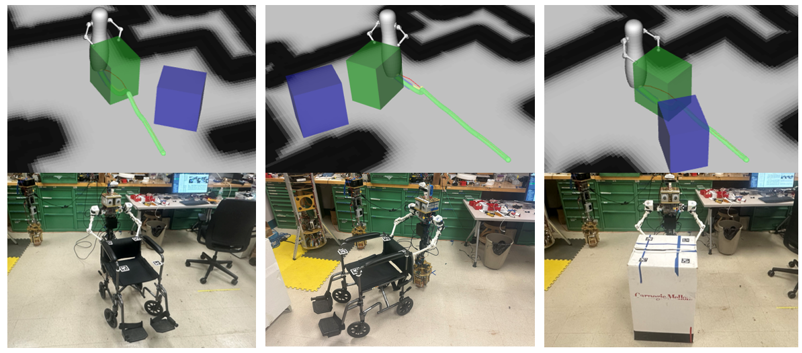}
    \caption{\textbf{Object Planner:} We show three planned object goals (blue) for the unregistered obstructing object (green), corresponding to a wheelchair with the right wheel locked, a cart with the front-left wheel locked, and a box. The green line is the global path planned by the collision-free navigation planner, the green box is the object's current position and the blue box is the desired goal planner by the object planner.}
    \label{fig:objectplanner}
    \centering
    \vspace*{-3mm}
\end{figure}

\subsubsection{Robot Dynamics}
To learn the dynamics model for the robot, we used supervised learning via stochastic gradient descent (SGD) to train the MLP, where the learning loss $\mathcal{L}_r$ formulated is: 
\begin{equation}
    \mathcal{L}_r=\frac{1}{n} \sum_{i=1}^{n} \left \| y^r - R(\mathbf{x}_t^r,\mathbf{u}_t) \right \|^2, 
\end{equation}
where $y^r$ is the ground truth value of robot acceleration and interaction force.

\subsubsection{Online Deployment}
 To update the linear coefficients online, we used the same update law from DAIML:
\begin{equation}
    \begin{aligned}
\mathbf{\dot{\hat{a}}}(c) & =-\lambda \mathbf{\hat{a}}(c) -P \psi^{\top} R^{-1}(\psi \mathbf{\hat{a}}(c)-\mathbf{y}^o) +P \psi^{\top} s,\\
\dot{P} & =-2 \lambda P+Q-P \psi^{\top} R^{-1} \psi P,
\end{aligned}
\end{equation}
where $\mathbf{\hat{a}}(c)$ is the online estimation of $\hat{a}(c)$, and $\dot{\hat{a}}$ is the online linear coefficient update used as $
\mathbf{\hat{a}}(c) = \mathbf{\hat{a}}(c) + \mathbf{\dot{\hat{a}}}(c)$; $P$ is a covariance-like matrix used for automatic gain tuning; $y^o$ is the measured object acceleration and $K, \Lambda,R,Q$ and $\lambda$ are gains.

To bridge the sim-to-real gap of the robot model between simulation and the real world, we fine-tuned $R$ with 5 minutes of real-world data $\mathcal{D}_r$. The data is collected by teleoperating the robot to interact with the environment continuously and record the time-stamped states where $\mathcal{D}_r=\left\{\left(\mathbf{x}^r_t, \mathbf{\dot{x}}^r_t, \mathbf{u}_t, \Gamma^r_t\right)\mid t=1, \ldots, T\right\}$. Since the 3-DoF arms are lightweight, we directly estimate $\Gamma^r_t$ by $\Gamma^r_t = J\tau_t$, where $J$ is the arm Jacobian matrix and $\tau_t$ is the joint torque.

 \begin{figure*}[t!]
    \centering
    \includegraphics[width=2\columnwidth]{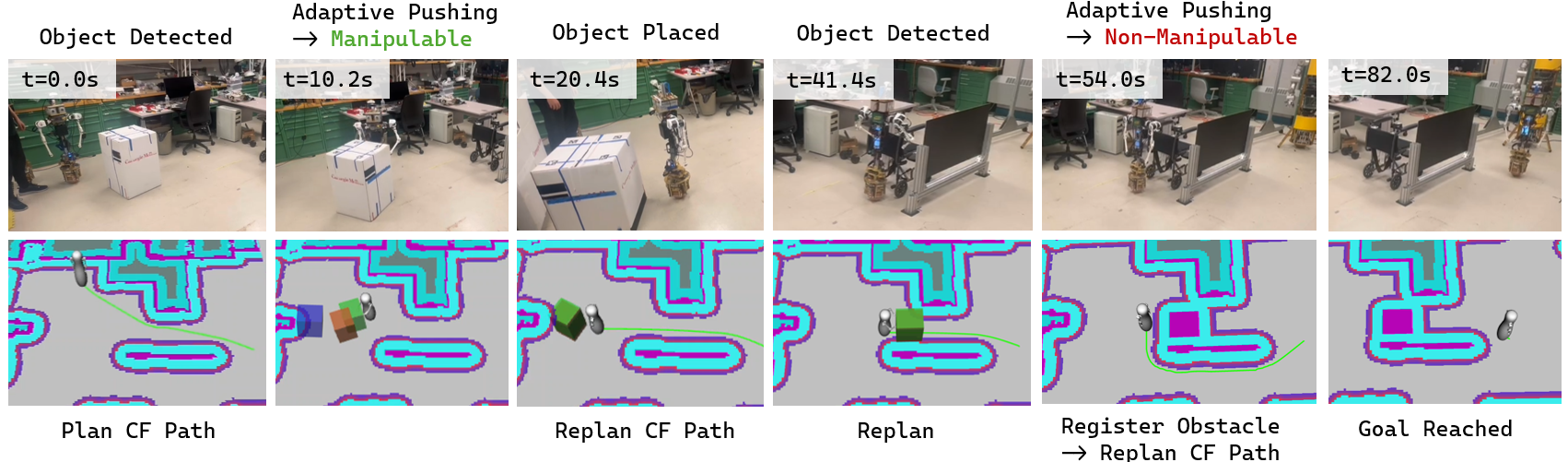}
    \caption{\textbf{Interactive navigation:} The top row shows time-stamped pictures of the interactive navigation experiment, where the top row labels illustrate the object-detection stage, and the bottom row illustrates the plan for collision-free path (CF Path). The bottom row is the corresponding visualization in Rviz.
 }
    \label{fig:finaldemo}
    \centering
    \vspace*{-3mm}
\end{figure*}

\begin{figure}[t!]
    \centering
    \includegraphics[width=\columnwidth]{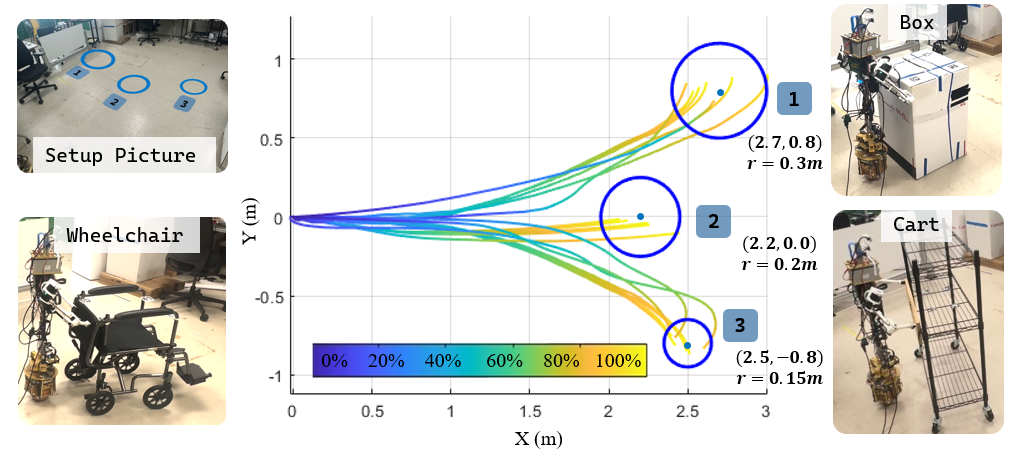}
    \caption{\textbf{Pushing Experiment with Movable Objects:} We evaluated the pushing performance across different objects and dynamics. The recorded object trajectories during adaptive pushing are shown with three goal locations, indicated in the top-left figure. The pushing progress is visualized with different colors representing the progress of the trajectory execution. Additionally, the graph highlights the goal locations and the minimum circle radius that encompasses all final object positions. }
    \label{fig:realpush}
    \centering
    \vspace*{-3mm}
\end{figure}

\begin{figure}[t!]
    \centering
    \includegraphics[width=\columnwidth]{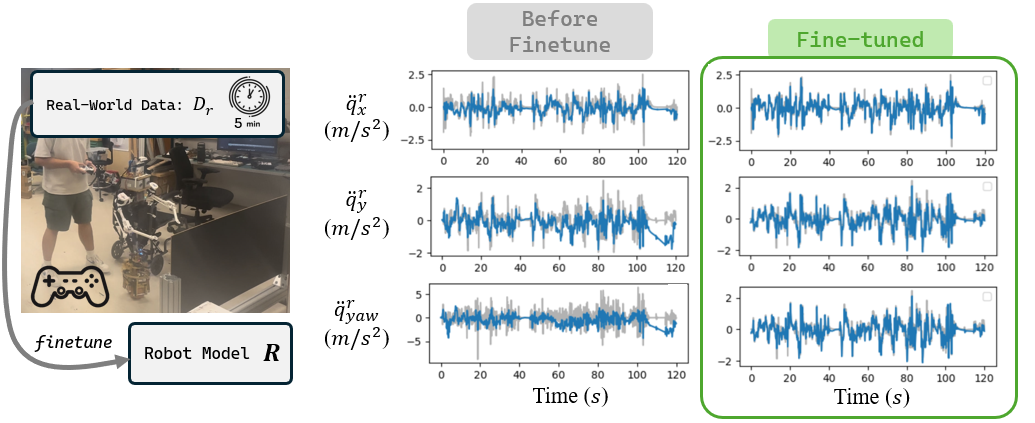}
    \caption{\textbf{Fine-tune Robot Model in Real World:} We fine-tuned the robot model $R$ using 5 minutes of real-world data collected via teleoperation, demonstrating a $27.3$ reduction in prediction MSE on a separate 2-minute real-world dataset. }
    \label{fig:finetune}
    \centering
    \vspace*{-3mm}
\end{figure}

\subsection{MPPI}
To plan for non-prehensile object manipulation, we leverage MPPI~\cite{7989202} with the learned adaptive model to predict and roll out $S$ possible trajectories with Length $T$, and then execute the first step of it.  The MPPI controller runs at 10 Hz with $S=500$ and $T = 30$, which corresponds to approximately 3~s of prediction. The temperature was set as $\lambda=0.7$ and the cost function $C_t(\mathbf{x,u}) $~is the Euclidean distance to the goal with a smoothing term that penalizes aggressive changes:
\begin{equation}
\begin{split}
C_t(\mathbf{x}, \mathbf{u}) = \|  \mathbf{x}^{o,des}_{[x,y,yaw]}-\mathbf{x}^{o,T}_{[x,y,yaw]} \|_2 \\+\sum_{k=1}^{T} ( \lambda_1 \| \mathbf{u}_k \|_2 
+ \lambda_2 \| \mathbf{u}_k - \mathbf{u}_{k-1} \|_2 ),
\end{split}
\end{equation}where $\mathbf{x}^{o,T}_{[x,y,yaw]}$ is the last state of the trajectory and $\mathbf{x}^{o,des}_{[x,y,yaw]}$ is the current and desired pose of the object, $\lambda_1$ and $\lambda_2$ are the penalizing coefficient for large action magnitude and change respectively.

\section{Interactive Navigation \label{sec:InteractiveNavigation}}
 Our goal is to enable the robot to reposition obstructing objects to make a path for collision-free navigation. To achieve this, the robot needs to further reason on whether the object is manipulable. We address this by introducing an object planner to the framework as illustrated in Fig.~\ref{fig:ineractive navigation}.

\subsection{Collision-Free Navigation}
The global path planner utilizes an $A^*$ algorithm~\cite{tbd} to find a global optimal path given start and goal positions. Since the robot is accelerated by its lean angle $\phi$, the local plan needs to refine the trajectory to be feasible for a ball-balancing system. This is done by solving a differentially-flat trajectory optimization given local goals and dynamical constraints of the robot~\cite{6631211}.

\subsection{Object Planner}
\begin{algorithm}
\label{alg:objectplanner}
\caption{Object Planner}
\SetKwInOut{Input}{input}\SetKwInOut{Output}{output}
    \Input{$M$: Map; \quad $P$: CF-Path; \\
    $\textbf{x}^{o}_{t}$: Current object state; \\}
    \Output{$g^{*}$: Optimal object goal }
    \textbf{Initilization Stage ($0<t\le 1$): }\\
    ${G}=empty GoalBuffer$ \\
    \For{$g\in $} {
        $g \gets CollisionCheck(M(2m \times 2m),P,g)$\\
        append $g$ to $G$
    }
    $g^*=ClosestGoal(G,\textbf{x}^{o}_{t})$ \\
    \textbf{Planning Stage ($t>1$):}\\
    \eIf{$K= \sum_{n=1}^{3} ( \| \textbf{x}^{*,T}_{[x,y]} - \textbf{x}^o_{t,[x,y]} \|^2  )<0.3$}
    {Register as obstacle on $M$, replan CF-path}
    {Sample rollout actions ${U}^{1\cdots N}_{1\cdots T}$\\
    \For{$x^{o,T}\in DynamicsRollout(\textbf{x}^{o}_{t}, {U}^{1\cdots N}_{1\cdots T})$}{
        $g \gets CollisionCheck(M,P,x^{o,T})$\\
        append $g$ to $G$\\
    }
    $g^*=ClosestGoal(G, \textbf{x}^{o}_{t})$ \\
    }
    \KwResult{$g*$}
\end{algorithm}
The object planner generates an optimal goal position $g^*$ that is kino-dynamically feasible for the manipulated object while identifying non-manipulable objects, such as heavy items or shelves fixed to the ground. First, we select a collision-free goal within a $2m \times 2m$ area centered at $\mathbf{x}^{o}_{t,[x,y]}$ to initialize the adaptation process. After 1~s of adaptation, we leverage the MPPI controller's dynamics roll-out infrastructure and do the same collision check over the last state of 500 trajectories with $T=50$. Figure~\ref{fig:ineractive navigation} illustrates several examples of the planned object goals. To identify the non-manipulable objects, we leverage the criteria $G$ which is the Euclidean distance between the current object pose $\mathbf{x}^{o}_{t,[x,y]}$ and the average object pose over the $n$ optimal roll-out trajectories at the end $\mathbf{x}^{*, T}_{[x,y]}$. The details of the object planner are shown in Algorithm 1.

\section{Method Validation\label{sec:Validation}}
\subsection{Object Model Accuracy}
We validate the expressiveness of the proposed object model by testing its prediction over 20 different objects not seen during training. In our setting, $\psi$ has four fully-connected hidden layers as $[10, 128, 128, 50, 6]$, and $R$ also has four fully-connected hidden layers as $[28, 128, 128, 50, 5]$. First, we use $25s$ of data to calculate $\mathbf{a}^*$, then use $\mathbf{a}^*$ to calculate prediction as shown in Eq.\ref{eq:psia}. The average Mean-Squared Error (MSE) of $O$ is $1.3131$ before learning and $0.4271$ after learning. We selected 8 different objects to show the prediction result in Fig.~\ref{fig:modelprediction}, 

\subsection{Baseline Study}
We further compared our formulation of $O$ with two baselines: \textbf{trigonometric $\psi$} and \textbf{residual adaptive $\psi$}. 

\subsubsection{\textbf{Trigonometric $\psi$ baseline}}
We selected a set of trigonometric functions to be the representation basis function $\psi$. The specific set of time-dependent trigonometric function set we selected is $\left[\sin(t), \sin(2t), \sin(3t), \cos(t), \cos(2t), \cos(3t) \right]^{\top}$. The mean MSE baseline across 20 testing environments is $0.5720$ which is higher compared to our method by $0.145$.

\subsubsection{\textbf{Adaptive Residual $\psi$}}
We first train an MLP, $N$, via supervised learning across $D_c$, then train $\psi$ with the same method with the goal of only adapting to the residual $r = \mathbf{y}_c-N(\mathbf{x}_t^o,\Gamma^o)$. The mean MSE in the adaptive residual $\psi$ baseline across 20 testing environments $0.4820$, which is only slightly higher than our method and outperforms our method on 2 objects. 

\section{Experiment\label{sec:Experiment}}
In this section, we validated the adaptive pushing controller and interactive navigation framework with real-world experiments.
\subsection{Hardware Platform and Experiment Setup}
 We used an RGB-Depth camera (Intel RealSense D435i) and a tracking camera (Intel Realsense T265) placed on top of Shmoobot to estimate ego-centric object pose and pose estimation. We attached four AprilTags~\cite{5979561} to the object's corners for pose estimation and size. We fine-tuned $R$ before running real-world experiments as discussed in Sec.\ref{sec:AdaptivePushing}, the results are shown in Fig.\ref{fig:finetune}.

\subsubsection{Pushing Movable Objects}
To demonstrate the effectiveness of the adaptive pushing controller, we experimented with three types of objects with Shmoobot. We changed object dynamics randomly during experiments such as: putting additional weight on the item and adding friction on the cart and wheelchair wheels by pressing the wheel brakes. We set three goal positions that are dynamically feasible for the objects as specified in Fig.\ref{fig:realpush} and randomly selected the goals during the experiment. The trajectories of the objects are plotted in Fig.~\ref{fig:realpush}, with he average distance offset is 0.067~m for Goal~1, 0.124~m for Goal~2, and 0.164~m for Goal~3. We also calculated the circle centered at each goal that contains all the final object positions shown in Fig.~\ref{fig:realpush}.

\subsection{Interactive Navigation Among Movable Objects}

In this experiment, we assembled all the components and validated the effectiveness of the proposed interactive navigation framework. We set two obstructing objects: a manipulable box (mass = 2.8~kg) and a non-manipulable wheelchair (mass = 11.9~kg, wheels locked). A time-elapse picture of this experiment is shown in 
Fig.~\ref{fig:introduction}, and the time-stamped pictures are shown in Fig.~\ref{fig:finaldemo} along with status illustration. This demonstrates the effectiveness of the proposed framework deployed on a dynamic mobile robot.

\section{Conclusion and Discussions\label{sec:Conclusion}}
This paper presents an interactive navigation framework with adaptive non-prehensile manipulation. The framework leverages learned representations of common indoor movable objects, which are validated through both simulation and hardware experiments. Future research could focus on adapting to more complex tasks, such as door-opening.





%

\newpage
\addtolength{\textheight}{-9cm} 
\addtolength{\textheight}{+10cm}
\bibliographystyle{IEEEtran}
\bibliography{IEEEabrv,ref.bib}

\begin{thebibliography}{10}
\providecommand{\url}[1]{#1}
\csname url@samestyle\endcsname
\providecommand{\newblock}{\relax}
\providecommand{\bibinfo}[2]{#2}
\providecommand{\BIBentrySTDinterwordspacing}{\spaceskip=0pt\relax}
\providecommand{\BIBentryALTinterwordstretchfactor}{4}
\providecommand{\BIBentryALTinterwordspacing}{\spaceskip=\fontdimen2\font plus
\BIBentryALTinterwordstretchfactor\fontdimen3\font minus \fontdimen4\font\relax}
\providecommand{\BIBforeignlanguage}[2]{{%
\expandafter\ifx\csname l@#1\endcsname\relax
\typeout{** WARNING: IEEEtran.bst: No hyphenation pattern has been}%
\typeout{** loaded for the language `#1'. Using the pattern for}%
\typeout{** the default language instead.}%
\else
\language=\csname l@#1\endcsname
\fi
#2}}
\providecommand{\BIBdecl}{\relax}
\BIBdecl

\bibitem{6224884}
S.~Nozawa, Y.~Kakiuchi, K.~Okada, and M.~Inaba, ``Controlling the planar motion of a heavy object by pushing with a humanoid robot using dual-arm force control,'' in \emph{2012 IEEE International Conference on Robotics and Automation}, 2012, pp. 1428--1435.

\bibitem{6224970}
U.~Nagarajan, G.~Kantor, and R.~Hollis, ``Integrated planning and control for graceful navigation of shape-accelerated underactuated balancing mobile robots,'' in \emph{2012 IEEE International Conference on Robotics and Automation}, 2012, pp. 136--141.

\bibitem{10160507}
A.~Rigo, Y.~Chen, S.~K. Gupta, and Q.~Nguyen, ``Contact optimization for non-prehensile loco-manipulation via hierarchical model predictive control,'' in \emph{2023 IEEE International Conference on Robotics and Automation (ICRA)}, 2023, pp. 9945--9951.

\bibitem{7989202}
G.~Williams, N.~Wagener, B.~Goldfain, P.~Drews, J.~M. Rehg, B.~Boots, and E.~A. Theodorou, ``Information theoretic mpc for model-based reinforcement learning,'' in \emph{2017 IEEE International Conference on Robotics and Automation (ICRA)}, 2017, pp. 1714--1721.

\bibitem{coumans2016pybullet}
E.~Coumans and Y.~Bai, ``Pybullet, a python module for physics simulation for games, robotics and machine learning,'' 2016.

\bibitem{9145591}
J.~Pankert and M.~Hutter, ``Perceptive model predictive control for continuous mobile manipulation,'' \emph{IEEE Robotics and Automation Letters}, vol.~5, no.~4, pp. 6177--6184, 2020.

\bibitem{Mericli}
T.~Mericli, M.~Veloso, and H.~L. Akin, ``Push-manipulation of complex passive mobile objects using experimentally acquired motion models,'' \emph{Autonomous Robots}, vol.~38, 03 2014.

\bibitem{9387121}
J.-P. Sleiman, F.~Farshidian, M.~V. Minniti, and M.~Hutter, ``A unified mpc framework for whole-body dynamic locomotion and manipulation,'' \emph{IEEE Robotics and Automation Letters}, vol.~6, no.~3, pp. 4688--4695, 2021.

\bibitem{ferrolho2023roloma}
H.~Ferrolho, V.~Ivan, W.~Merkt, I.~Havoutis, and S.~Vijayakumar, ``Roloma: Robust loco-manipulation for quadruped robots with arms,'' \emph{Autonomous Robots}, vol.~47, no.~8, pp. 1463--1481, 2023.

\bibitem{9034999}
R.~Shu and R.~Hollis, ``Development of a humanoid dual arm system for a single spherical wheeled balancing mobile robot,'' in \emph{2019 IEEE-RAS 19th International Conference on Humanoid Robots (Humanoids)}, 2019, pp. 499--504.

\bibitem{li2023kinodynamicsbasedposeoptimizationhumanoid}
\BIBentryALTinterwordspacing
J.~Li and Q.~Nguyen, ``Kinodynamics-based pose optimization for humanoid loco-manipulation,'' 2023. [Online]. Available: \url{https://arxiv.org/abs/2303.04985}
\BIBentrySTDinterwordspacing

\bibitem{8758922}
M.~V. Minniti, F.~Farshidian, R.~Grandia, and M.~Hutter, ``Whole-body mpc for a dynamically stable mobile manipulator,'' \emph{IEEE Robotics and Automation Letters}, vol.~4, no.~4, pp. 3687--3694, 2019.

\bibitem{aguilera_control_2023}
\BIBentryALTinterwordspacing
S.~Aguilera and S.~Hutchinson, ``Control of cart-like nonholonomic systems using a mobile manipulator,'' in \emph{2023 {IEEE}/{RSJ} International Conference on Intelligent Robots and Systems ({IROS})}.\hskip 1em plus 0.5em minus 0.4em\relax {IEEE}, 2023, pp. 6801--6808. [Online]. Available: \url{https://ieeexplore.ieee.org/document/10342088/}
\BIBentrySTDinterwordspacing

\bibitem{dai2024wheelchairmaneuveringsinglesphericalwheeledbalancing}
\BIBentryALTinterwordspacing
C.~Dai, X.~Liu, R.~Shu, and R.~Hollis, ``Wheelchair maneuvering with a single-spherical-wheeled balancing mobile manipulator,'' 2024. [Online]. Available: \url{https://arxiv.org/abs/2404.13206}
\BIBentrySTDinterwordspacing

\bibitem{kumar2021rma}
A.~Kumar, Z.~Fu, D.~Pathak, and J.~Malik, ``Rma: Rapid motor adaptation for legged robots,'' in \emph{Robotics: Science and Systems}, 2021.

\bibitem{li2023robust}
Z.~Li, X.~B. Peng, P.~Abbeel, S.~Levine, G.~Berseth, and K.~Sreenath, ``Robust and versatile bipedal jumping control through reinforcement learning,'' \emph{arXiv preprint arXiv:2302.09450}, 2023.

\bibitem{pmlr-v70-finn17a}
\BIBentryALTinterwordspacing
C.~Finn, P.~Abbeel, and S.~Levine, ``Model-agnostic meta-learning for fast adaptation of deep networks,'' in \emph{Proceedings of the 34th International Conference on Machine Learning}, ser. Proceedings of Machine Learning Research, D.~Precup and Y.~W. Teh, Eds., vol.~70.\hskip 1em plus 0.5em minus 0.4em\relax PMLR, 06--11 Aug 2017, pp. 1126--1135. [Online]. Available: \url{https://proceedings.mlr.press/v70/finn17a.html}
\BIBentrySTDinterwordspacing

\bibitem{Kopicki}
M.~Kopicki, S.~Zurek, R.~Stolkin, T.~Mörwald, and J.~Wyatt, ``Learning modular and transferable forward models of the motions of push manipulated objects,'' \emph{Autonomous Robots}, vol.~41, 06 2017.

\bibitem{O_Connell_2022}
\BIBentryALTinterwordspacing
M.~O’Connell, G.~Shi, X.~Shi, K.~Azizzadenesheli, A.~Anandkumar, Y.~Yue, and S.-J. Chung, ``Neural-fly enables rapid learning for agile flight in strong winds,'' \emph{Science Robotics}, vol.~7, no.~66, May 2022. [Online]. Available: \url{http://dx.doi.org/10.1126/scirobotics.abm6597}
\BIBentrySTDinterwordspacing

\bibitem{6631116}
M.~Levihn, J.~Scholz, and M.~Stilman, ``Planning with movable obstacles in continuous environments with uncertain dynamics,'' in \emph{2013 IEEE International Conference on Robotics and Automation}, 2013, pp. 3832--3838.

\bibitem{7759546}
J.~Scholz, N.~Jindal, M.~Levihn, C.~L. Isbell, and H.~I. Christensen, ``Navigation among movable obstacles with learned dynamic constraints,'' in \emph{2016 IEEE/RSJ International Conference on Intelligent Robots and Systems (IROS)}, 2016, pp. 3706--3713.

\bibitem{schoch2024insightinteractivenavigationsight}
\BIBentryALTinterwordspacing
P.~Schoch, F.~Yang, Y.~Ma, S.~Leutenegger, M.~Hutter, and Q.~Leboutet, ``In-sight: Interactive navigation through sight,'' 2024. [Online]. Available: \url{https://arxiv.org/abs/2408.00343}
\BIBentrySTDinterwordspacing

\bibitem{8954627}
F.~Xia, W.~B. Shen, C.~Li, P.~Kasimbeg, M.~E. Tchapmi, A.~Toshev, R.~Martín-Martín, and S.~Savarese, ``Interactive gibson benchmark: A benchmark for interactive navigation in cluttered environments,'' \emph{IEEE Robotics and Automation Letters}, vol.~5, no.~2, pp. 713--720, 2020.

\bibitem{khz2021interact}
K.-H. Zeng, A.~Farhadi, L.~Weihs, and R.~Mottaghi, ``Pushing it out of the way: Interactive visual navigation,'' in \emph{CVPR}, 2021.

\bibitem{he2024interactivefarinteractivefastadaptablerouting}
\BIBentryALTinterwordspacing
B.~He, G.~Chen, W.~Wang, J.~Zhang, C.~Fermuller, and Y.~Aloimonos, ``Interactive-far:interactive, fast and adaptable routing for navigation among movable obstacles in complex unknown environments,'' 2024. [Online]. Available: \url{https://arxiv.org/abs/2404.07447}
\BIBentrySTDinterwordspacing

\bibitem{lauwers_dynamically_2006}
T.~Lauwers, G.~Kantor, and R.~Hollis, ``A dynamically stable single-wheeled mobile robot with inverse mouse-ball drive,'' in \emph{Proceedings 2006 {IEEE} {International} {Conference} on {Robotics} and {Automation}, 2006. {ICRA} 2006.}, May 2006, pp. 2884--2889, iSSN: 1050-4729.

\bibitem{srivatchan_development_nodate}
S.~Srivatchan, ``Development of a balancing robot as an indoor service agent,'' 2020.

\bibitem{6225065}
U.~Nagarajan, B.~Kim, and R.~Hollis, ``Planning in high-dimensional shape space for a single-wheeled balancing mobile robot with arms,'' in \emph{2012 IEEE International Conference on Robotics and Automation}, 2012, pp. 130--135.

\bibitem{lutter2021learningdynamicsmodelsmodel}
\BIBentryALTinterwordspacing
M.~Lutter, L.~Hasenclever, A.~Byravan, G.~Dulac-Arnold, P.~Trochim, N.~Heess, J.~Merel, and Y.~Tassa, ``Learning dynamics models for model predictive agents,'' 2021. [Online]. Available: \url{https://arxiv.org/abs/2109.14311}
\BIBentrySTDinterwordspacing

\bibitem{hafner2019learninglatentdynamicsplanning}
\BIBentryALTinterwordspacing
D.~Hafner, T.~Lillicrap, I.~Fischer, R.~Villegas, D.~Ha, H.~Lee, and J.~Davidson, ``Learning latent dynamics for planning from pixels,'' 2019. [Online]. Available: \url{https://arxiv.org/abs/1811.04551}
\BIBentrySTDinterwordspacing

\bibitem{Venkatraman2015ImprovingMP}
\BIBentryALTinterwordspacing
A.~Venkatraman, M.~Hebert, and J.~A. Bagnell, ``Improving multi-step prediction of learned time series models,'' in \emph{AAAI Conference on Artificial Intelligence}, 2015. [Online]. Available: \url{https://api.semanticscholar.org/CorpusID:15080650}
\BIBentrySTDinterwordspacing

\bibitem{tbd}
E.~Marder-Eppstein, E.~Berger, T.~Foote, B.~Gerkey, and K.~Konolige, ``The office marathon: Robust navigation in an indoor office environment,'' in \emph{International Conference on Robotics and Automation}, 2010.

\bibitem{6631211}
M.~Shomin and R.~Hollis, ``Differentially flat trajectory generation for a dynamically stable mobile robot,'' in \emph{2013 IEEE International Conference on Robotics and Automation}, 2013, pp. 4467--4472.

\bibitem{5979561}
E.~Olson, ``Apriltag: A robust and flexible visual fiducial system,'' in \emph{2011 IEEE International Conference on Robotics and Automation}, 2011, pp. 3400--3407.

\end{thebibliography}

\end{document}